\documentclass[letterpaper]{article} 
\usepackage[preprint]{aaai2027}  
\usepackage[hyphens]{url}  
\usepackage{graphicx} 
\usepackage{natbib}  
\usepackage{caption} 
\usepackage{algorithm}
\usepackage{algorithmic}

\usepackage{newfloat}
\usepackage{listings}
\DeclareCaptionStyle{ruled}{labelfont=normalfont,labelsep=colon,strut=off} 
\floatstyle{ruled}
\newfloat{listing}{tb}{lst}{}
\floatname{listing}{Listing}

\usepackage{booktabs}

\usepackage{amssymb}
\usepackage{amsmath}

\title{FedLAFP: Low-Rank Aggregation Meets Full-Rank Personalization \\in Federated Fine-Tuning}
\author{
    Mengjun Yi\textsuperscript{\rm 1,\rm 2},
    Huaian Gu\textsuperscript{\rm 1,\rm 2},
    Yinghao Ai\textsuperscript{\rm 1,\rm 3},
    Furao Shen\textsuperscript{\rm 1,\rm 2}\corresponding,
    Jian Zhao\textsuperscript{\rm 4}
}
\affiliations{
    \textsuperscript{\rm 1}State Key Laboratory for Novel Software Technology\\
    \textsuperscript{\rm 2}School of Artificial Intelligence\\
    \textsuperscript{\rm 3}School of Computer Science\\
    \textsuperscript{\rm 4}School of Electronic Science and Engineering\\

    Nanjing University\\
    Nanjing, 210023, China\\
    mengjunyi@smail.nju.edu.cn, frshen@nju.edu.cn
}

\begin{document}

\maketitle

\begin{abstract}
Federated parameter-efficient fine-tuning enables clients to adapt pre-trained models without sharing raw data or communicating the full model, but statistical heterogeneity makes a single global adapter insufficient for personalized prediction.
Existing personalized methods typically use the same low-rank structure for both shared and private adaptation, overlooking their distinct requirements for aggregation and personalization.
We propose FedLAFP, a role-aware framework that couples a compact, globally aggregated LoRA branch with a client-private, full-rank-capable RandLoRA branch.
The shared branch provides an efficient interface for transferring common knowledge, whereas the private branch combines fixed random low-rank bases with learned scaling coefficients to provide expressive client-specific adaptation without additional communication.
Client- and layer-specific mixing coefficients jointly fuse the two branches, and only the shared LoRA parameters are exchanged.
A controlled linear study supports this role assignment: LoRA yields more aligned client updates and lower aggregation error, while RandLoRA more accurately recovers client-specific residuals.
Experiments across four visual recognition benchmarks show that FedLAFP consistently outperforms local-only and federated LoRA baselines, achieving an average personalized accuracy of $86.93\%$ and exceeding the best baseline average by $1.30$ percentage points.
\end{abstract}


\section{Introduction}
\label{sec:introduction}

Adapting pre-trained foundation models often relies on task data distributed across users, institutions, or edge devices~\cite{guo2023promptfl}. 
Federated learning (FL)~\cite{mcmahan2017communication} allows these clients to train collaboratively without centralizing raw data, but optimizing and repeatedly communicating an entire foundation model is expensive.
Parameter-efficient fine-tuning (PEFT)~\cite{fu2023effectiveness}, particularly low-rank adaptation (LoRA)~\cite{hu2022lora}, mitigates this cost by freezing the backbone and learning compact low-rank updates.

\begin{figure}[t]
    \centering
    \includegraphics[width=\columnwidth]{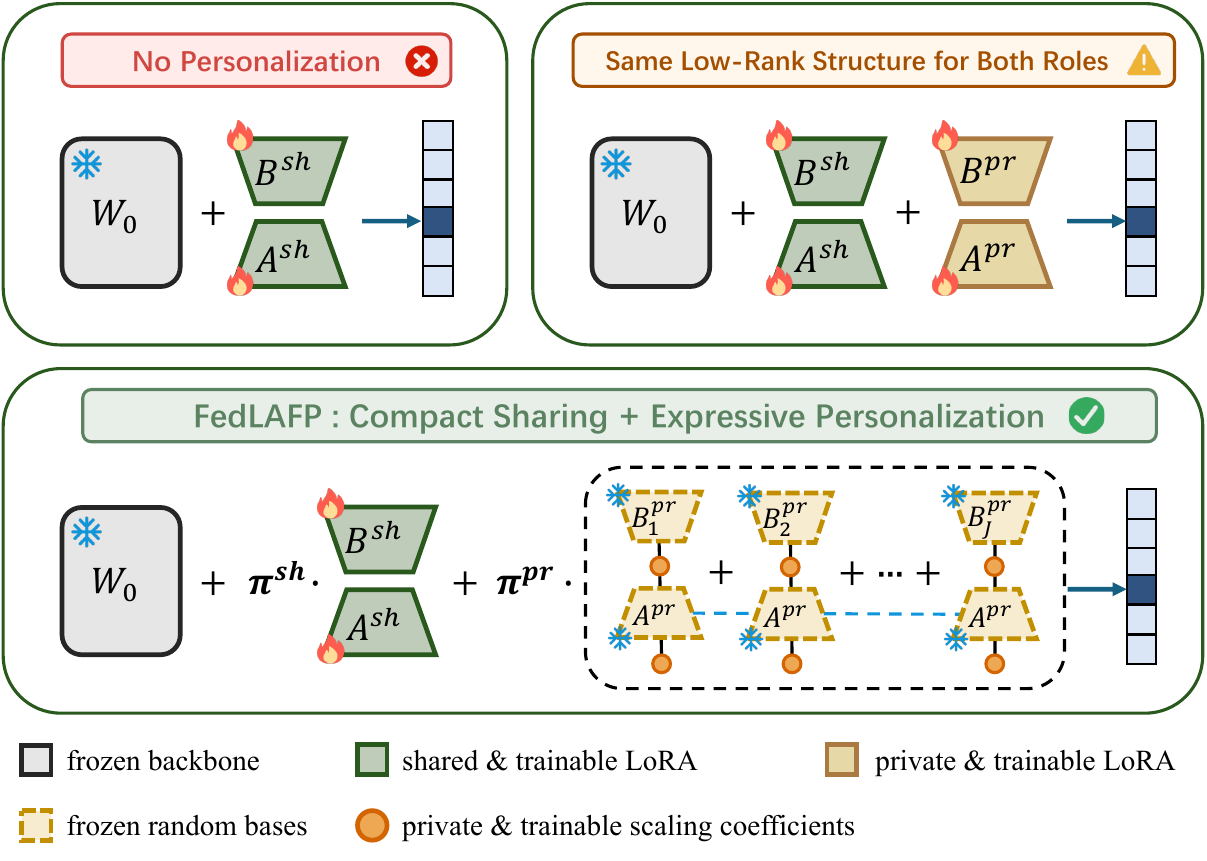}
    \caption{Comparison of global-only adaptation, symmetric shared--private LoRA, and the proposed role-aware FedLAFP.}
    \label{fig:lora-paradigms}
\end{figure}

However, statistical heterogeneity remains a key challenge for federated LoRA~\cite{chen2026fedmerge}.
Across heterogeneous clients, a single global LoRA may dilute client-specific information and induce interference, whereas independently trained local adapters forgo cross-client knowledge sharing~\cite{yang2025federated}.
Personalized federated PEFT balances these objectives through shared and private components~\cite{bian2026fedalt}. 
Although effective, most such designs use the same adapter family for both components, typically instantiating each as a structurally identical LoRA module~\cite{yang2024dual,lu2024fdlora,bian2026fedalt}. 
As illustrated in Figure~\ref{fig:lora-paradigms}, this symmetric design overlooks the distinct roles of shared and private adaptation.

Although clients have different local distributions, their data usually contain both common patterns shared across clients and characteristics specific to each client.
Personalized adaptation therefore serves two distinct roles: the shared component captures knowledge that generalizes across clients, whereas the private component models the characteristics of each client's local distribution.
These roles lead to different design requirements.
Because the shared component is repeatedly communicated and aggregated, it should be compact and aggregation-friendly.
The private component remains local and is not averaged across clients, so it should have sufficient capacity to capture client-specific variations.
Applying the same low-rank bottleneck to both components may therefore restrict personalization, motivating a role-dependent adapter design.

Motivated by these different requirements, we propose FedLAFP, a role-aware personalized federated adaptation framework that combines a shared LoRA branch with a client-private RandLoRA~\cite{albert2025randlora} branch.
LoRA factorizes the shared update into two trainable low-rank matrices, keeping the communicated parameter set compact.
RandLoRA constructs a full-rank-capable private update as a sum of fixed random low-rank bases, each modulated by learnable diagonal scaling matrices. This provides a broader update space for local adaptation without directly optimizing a full-sized update matrix.
Both branches are attached to a frozen pre-trained backbone and jointly optimized on each client's data.
The server aggregates only the shared LoRA parameters, while each client’s private RandLoRA parameters remain local.
Client- and layer-specific mixing coefficients adaptively combine the shared and private updates.
Together, the two branches support efficient cross-client knowledge sharing and expressive local personalization.
Our contributions are as follows:
\begin{itemize}
    \item We revisit the symmetric dual-LoRA design commonly adopted in personalized federated fine-tuning and identify distinct requirements for its shared and private branches.
    The shared branch should remain compact for efficient communication and aggregation, whereas the private branch requires greater expressive capacity to capture client-specific heterogeneity.

    \item We propose FedLAFP, a role-aware personalized federated fine-tuning framework that combines a globally aggregated LoRA branch with a client-private, full-rank-capable RandLoRA branch.
    Client- and layer-specific fusion weights adaptively balance the shared and private updates, while only the compact shared LoRA parameters are communicated.

    \item Experiments on four visual recognition benchmarks show that FedLAFP achieves the best accuracy on every dataset, outperforming local-only adaptation and a broad range of federated LoRA baselines covering global adapter aggregation, factor-wise selective sharing, and shared--private personalization.
\end{itemize}

\section{Related Work}
\label{sec:related-work}

\subsection{PEFT and LoRA Variants}

Parameter-efficient fine-tuning (PEFT) adapts a pre-trained model by optimizing only a small task-specific component while leaving most backbone parameters frozen~\cite{ding2023parameter}.
Representative approaches insert bottleneck adapters between Transformer layers~\cite{lu2023fedclip}, optimize continuous prefixes or soft prompts~\cite{zhou2022learning}, or update only selected parameters such as bias terms~\cite{zaken2022bitfit}.
Low-Rank Adaptation (LoRA)~\cite{hu2022lora} instead represents the update to a frozen weight matrix as the product of two trainable low-rank factors.
Its compact parameterization reduces communication overhead, while the learned update can be merged into the backbone without introducing additional inference latency. These properties make LoRA particularly attractive for federated learning.

LoRA variants extend the basic formulation along different dimensions.
AdaLoRA~\cite{zhang2023adalora} adaptively allocates rank budgets across weight matrices according to their importance scores, while DoRA~\cite{liu2024dora} decomposes pre-trained weights into magnitude and direction and applies LoRA to directional updates.
VeRA~\cite{kopiczko2024vera} shares frozen random low-rank matrices across layers and learns lightweight scaling vectors.
RandLoRA~\cite{albert2025randlora} learns combinations of fixed random low-rank matrices through diagonal scaling matrices, enabling full-rank-capable updates without directly optimizing a full-sized update matrix.

\subsection{Federated Learning and Personalization}

Federated learning enables collaborative model training without centralizing raw data, and FedAvg realizes this paradigm by alternating local optimization with sample-weighted aggregation.
However, under heterogeneous data distributions, a single global model may not adequately capture client-specific characteristics.
Personalized federated learning addresses this limitation by combining cross-client knowledge sharing with client-specific adaptation.

Personalized federated learning methods differ in when and how client-specific adaptation is introduced.
Some methods learn a shared initialization or representation that is subsequently adapted to each client: Per-FedAvg~\cite{fallah2020personalized} meta-learns a shared initialization that clients adapt with a few gradient steps, while FedBABU~\cite{oh2022fedbabu} federatively trains the feature extractor and later fine-tunes the prediction head locally.
Other methods retain private components throughout federated training while aggregating shared ones: FedRep~\cite{collins2021exploiting} alternates between private client heads and a shared representation, whereas Fed-RoD~\cite{chen2022on} combines a shared feature extractor and generic predictor with a lightweight personalized head.
Developed for conventional task-specific models rather than parameter-efficient adaptation of pre-trained models, these methods do not examine how different PEFT parameterizations should be assigned to shared and private roles.

\begin{figure*}[t]
    \centering
    \includegraphics[width=0.9\textwidth]{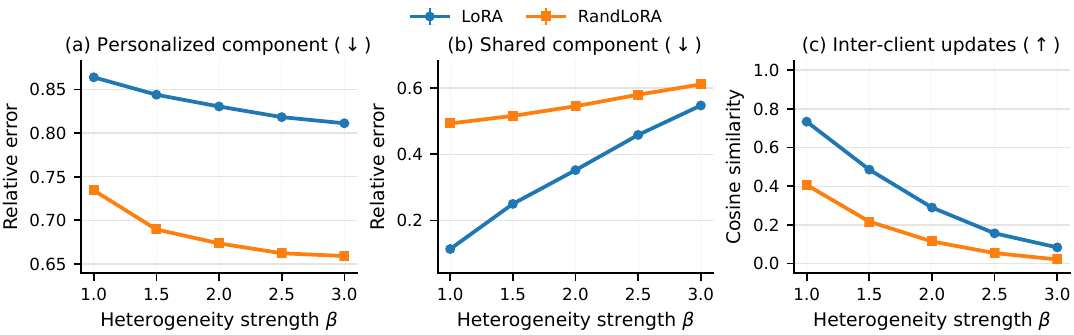}
    \caption{Controlled comparison of LoRA and RandLoRA under increasing client heterogeneity.
    Panels report (a) personalization error, (b) aggregation error, and (c) mean cross-client update similarity.}
    \label{fig:motivation}
\end{figure*}

\subsection{Federated LoRA Fine-Tuning}

Federated LoRA studies mainly explore global aggregation and personalization.
For global aggregation, FFA-LoRA~\cite{sunimproving} freezes one factor to avoid factor mismatch, while FedEx-LoRA~\cite{singhal2025fedex} corrects the residual caused by separately aggregating LoRA factors.
For personalization within a single adapter, FedSA-LoRA~\cite{guo2025selective} shares one factor and retains the other locally, whereas FedPissa~\cite{hefedpissa} separates shared and private subspaces.
Other methods maintain explicit shared and private branches: FedDPA~\cite{yang2024dual} and FDLoRA~\cite{lu2024fdlora} fuse global and local LoRA adapters, while FedALT~\cite{bian2026fedalt} combines an individual LoRA with a Rest-of-World LoRA through input-adaptive mixing.

Unlike these methods, which either separate shared and private knowledge within one low-rank adapter or use the same adapter family for both branches, FedLAFP assigns LoRA to shared aggregation and RandLoRA to private adaptation.

\section{Motivation}
\label{sec:motivation}

A natural question arising from the distinct requirements of shared and private adaptation is: \emph{which adapter is better suited to each federated role?}
To answer this question, we construct a controlled linear problem and examine two comparisons:
\textbf{Q1:} which adapter better captures client-specific variation during local adaptation?
\textbf{Q2:} which adapter better supports cross-client aggregation?

\subsubsection{Controlled setup.}
We construct a federated linear regression problem with $K$ clients and a frozen base model $W_0$.
The target update for client $i$ is
\begin{equation}
    \Delta W_i^\star
    = \Delta W_{\mathrm{sh}}^\star
    + \beta \Delta W_{i,\mathrm{pr}}^\star,
    \qquad
    \frac{1}{K}\sum_{i=1}^{K}\Delta W_{i,\mathrm{pr}}^\star = 0,
    \label{eq:synthetic-decomposition}
\end{equation}
where $\Delta W_{\mathrm{sh}}^\star$ is the shared component, $\Delta W_{i,\mathrm{pr}}^\star$ are the client-specific residuals, and $\beta$ controls the degree of heterogeneity.
Inputs are sampled from an isotropic Gaussian distribution, and targets are generated by applying $W_0+\Delta W_i^\star$ with additive Gaussian noise.
For each client, we train and evaluate LoRA and RandLoRA using exactly the same input--output samples, thereby isolating the effect of the adapter parameterization.

Because the synthetic construction provides known shared and private components, we can directly evaluate how well each adapter recovers them.
Let $\widehat{\Delta W}_i$ denote the update learned by client $i$ and $\overline{\Delta W}$ the average update across clients.
For personalization, we treat $\widehat{\Delta W}_i-\overline{\Delta W}$ as the recovered client-specific residual and measure its normalized Frobenius distance from the ground-truth private residual $\beta\Delta W_{i,\mathrm{pr}}^\star$; a lower error indicates better personalization.
For aggregation, we measure the normalized Frobenius distance between $\overline{\Delta W}$ and the ground-truth shared update $\Delta W_{\mathrm{sh}}^\star$.
We additionally compute the mean pairwise cosine similarity between client updates to quantify their directional consistency, where a lower aggregation error and a higher similarity indicate more aggregation-friendly updates.

\subsubsection{Q1: Which adapter better captures client-specific variation?}
Figure~\ref{fig:motivation}(a) shows that RandLoRA consistently achieves lower personalization error as heterogeneity increases.
By combining multiple fixed random low-rank bases, RandLoRA can represent a broader range of update directions and more accurately recover client-specific residuals.

\subsubsection{Q2: Which adapter better supports aggregation?}
Figures~\ref{fig:motivation}(b) and (c) show that LoRA yields lower aggregation error and more aligned client updates.
Although increasing heterogeneity makes aggregation more difficult for both adapters, LoRA consistently preserves shared directions more effectively.

Together, these results answer the opening question: RandLoRA is better suited to expressive private adaptation, whereas LoRA is better suited to compact shared aggregation.
This complementary role assignment directly motivates the asymmetric design of FedLAFP.

\section{Method}
\label{sec:method}

\begin{figure*}[t]
    \centering
    \includegraphics[width=\textwidth]{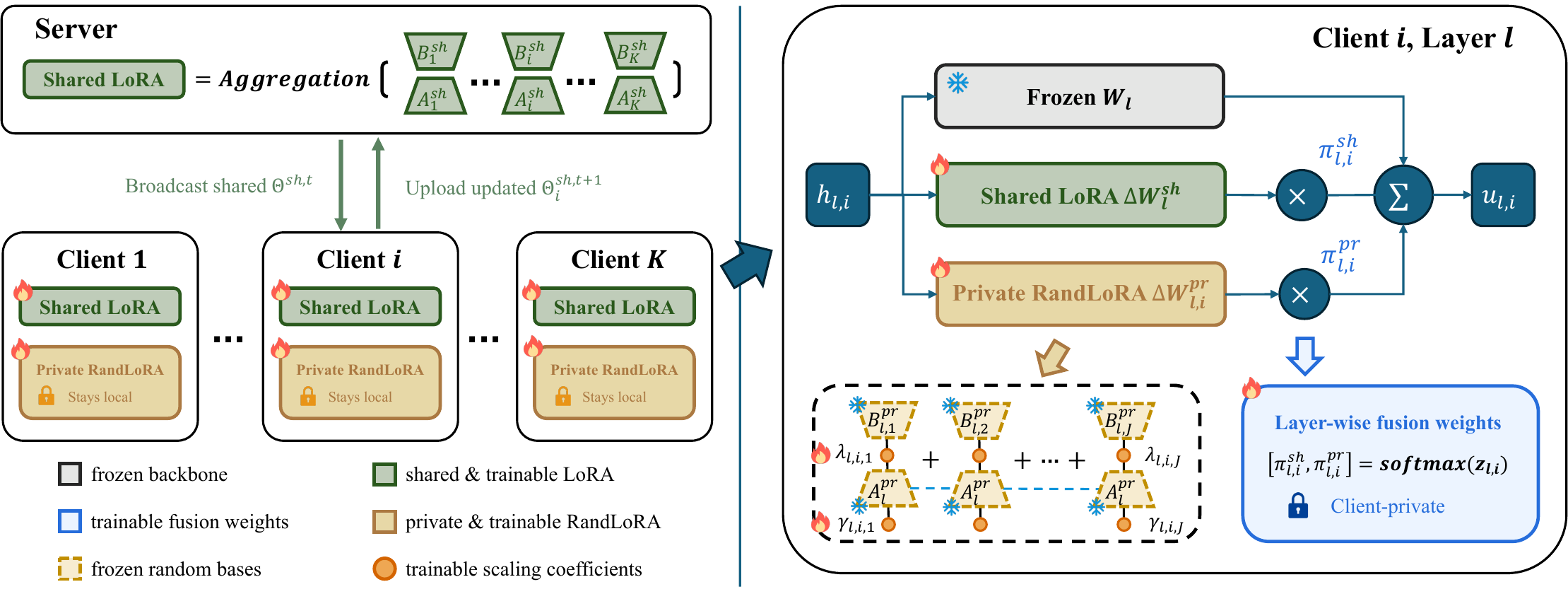}
    \caption{Overview of the FedLAFP framework. (a) The server broadcasts the shared LoRA parameters, and clients upload only their locally updated shared parameters for sample-weighted aggregation, while the private RandLoRA parameters remain local. (b) Within target layer $\ell$ of client $i$, the output of the frozen linear transformation is augmented by the weighted outputs of the shared LoRA and private RandLoRA branches, using $\pi_{\ell,i}^{\mathrm{sh}}$ and $\pi_{\ell,i}^{\mathrm{pr}}$, respectively. These client- and layer-specific fusion weights are derived from private logits and are never communicated.}
    \label{fig:framework}
\end{figure*}

Accordingly, FedLAFP uses a shared LoRA branch for cross-client aggregation and a private RandLoRA branch for local personalization.
As shown in Fig.~\ref{fig:framework}, the two branches are attached to each adapted layer of the frozen backbone and combined using client- and layer-specific weights.

\subsection{Problem Formulation}
\label{sec:method-formulation}

Consider $K$ clients, where client $i$ owns a local dataset $\mathcal{D}_i=\{(x_{i,n},y_{i,n})\}_{n=1}^{N_i}$ and $N=\sum_{i=1}^{K}N_i$.
Let $\Theta_0$ denote the parameters of a pre-trained backbone, which remain frozen throughout federated fine-tuning.
We introduce three groups of adaptation parameters.
The shared LoRA parameters $\Theta^{\mathrm{sh}}$ are synchronized through the server.
The RandLoRA parameters $\Theta_i^{\mathrm{pr}}$ and mixing logits $z_i$ are specific to client $i$ and persist locally across communication rounds.
The personalized model of client $i$ is therefore written as
\begin{equation}
    f_i(x)
    = f\!\left(
      x;
      \Theta_0,
      \Theta^{\mathrm{sh}},
      \Theta_i^{\mathrm{pr}},
      z_i
    \right).
    \label{eq:personalized-model}
\end{equation}
FedLAFP optimizes the sample-weighted personalized objective
\begin{equation}
    \min_{\Theta^{\mathrm{sh}},
          \{\Theta_i^{\mathrm{pr}},z_i\}_{i=1}^{K}}
    \sum_{i=1}^{K}\frac{N_i}{N}
    \mathcal{L}_i\!\left(
      \Theta^{\mathrm{sh}},
      \Theta_i^{\mathrm{pr}},z_i
    \right),
    \label{eq:fedlafp-objective}
\end{equation}
where
\begin{equation}
    \mathcal{L}_i
    = \frac{1}{N_i}
      \sum_{(x,y)\in\mathcal{D}_i}
      \ell\!\left(f_i(x),y\right).
    \label{eq:client-objective}
\end{equation}
Only $\Theta^{\mathrm{sh}}$ is communicated and aggregated by the server.
The private variables are optimized jointly with the shared branch during local training, but remain client-specific and are never aggregated.

\subsection{Role-Aware Heterogeneous Adaptation}
\label{sec:method-architecture}

Let $W_\ell\in\mathbb{R}^{d_{\mathrm{out}}\times d_{\mathrm{in}}}$ be a frozen weight matrix in a target layer $\ell$, and let $h_{\ell,i}$ denote its input on client $i$.
FedLAFP augments this matrix with one shared update and one private update.
The layer output is
\begin{equation}
    u_{\ell,i}
    = W_\ell h_{\ell,i}
    + \pi_{\ell,i}^{\mathrm{sh}}
      \Delta W_\ell^{\mathrm{sh}}h_{\ell,i}
    + \pi_{\ell,i}^{\mathrm{pr}}
      \Delta W_{\ell,i}^{\mathrm{pr}}h_{\ell,i}.
    \label{eq:dual-adapter-layer}
\end{equation}
The frozen projection and the two adapter updates are computed from the same input $h_{\ell,i}$ and summed to produce the layer output $u_{\ell,i}$.
In our experiments, we apply this construction to the query and value projection matrices in every Transformer attention block.
The same formulation can also be applied to other linear layers.

\subsubsection{Shared low-rank aggregation branch.}
The shared update is represented using LoRA:
\begin{equation}
    \Delta W_\ell^{\mathrm{sh}}
    = \frac{\alpha_L}{r_L}
      B_\ell^{\mathrm{sh}}A_\ell^{\mathrm{sh}},
    \label{eq:method-shared-lora}
\end{equation}
where $r_L$ is the LoRA rank, $\alpha_L$ is its scaling parameter, $A_\ell^{\mathrm{sh}}\in\mathbb{R}^{r_L\times d_{\mathrm{in}}}$, and $B_\ell^{\mathrm{sh}}\in\mathbb{R}^{d_{\mathrm{out}}\times r_L}$.
We initialize $A_\ell^{\mathrm{sh}}$ using Kaiming initialization and set $B_\ell^{\mathrm{sh}}$ to zero.
Since $B_\ell^{\mathrm{sh}}=0$, the shared update is initially zero, so the adapter does not alter the pre-trained model before fine-tuning.
As the only branch communicated and aggregated by the server, its low-rank factors provide a compact representation of knowledge shared across clients.

\subsubsection{Private full-rank-capable personalization branch.}
For the local branch, client $i$ uses RandLoRA.
Let $r_R$ denote the random-basis rank, $d_\ell=\min(d_{\mathrm{in}},d_{\mathrm{out}})$, and $J_\ell=\lceil d_\ell/r_R\rceil$.
The private update is
\begin{equation}
    \begin{aligned}
    \Delta W_{\ell,i}^{\mathrm{pr}}
    &= \frac{\alpha_R}{r_R}
      \sum_{j=1}^{J_\ell}
      B^{\mathrm{pr}}_{\ell,j}
      \operatorname{diag}(\lambda_{\ell,i,j})
      A^{\mathrm{pr}}_{\ell}
      \operatorname{diag}(\gamma_{\ell,i,j}),
    \end{aligned}
    \label{eq:method-private-randlora}
\end{equation}
where $\alpha_R$ is the RandLoRA scaling parameter, $A^{\mathrm{pr}}_{\ell}$ and $\{B^{\mathrm{pr}}_{\ell,j}\}_{j=1}^{J_\ell}$ are fixed random bases, and $\lambda_{\ell,i,j}$ and $\gamma_{\ell,i,j}$ are learned locally.
By summing multiple scaled low-rank components, RandLoRA is full-rank-capable and can capture richer client-specific deviations than a single low-rank update. The random bases are deterministically generated from a shared public seed, remain fixed during training, and require no communication. We initialize $\lambda_{\ell,i,j}$ to zero and $\gamma_{\ell,i,j}$ to a small constant so that the private update starts from zero. Only the client-specific scaling vectors are optimized and kept local.

\subsubsection{Client-specific layer-wise fusion.}
The two adapter branches need not contribute equally to every client or layer.
For each target layer, client $i$ maintains two trainable logits $z_{\ell,i}\in\mathbb{R}^{2}$ and computes
\begin{equation}
    \left[
      \pi_{\ell,i}^{\mathrm{sh}},
      \pi_{\ell,i}^{\mathrm{pr}}
    \right]
    = \operatorname{softmax}(z_{\ell,i}).
    \label{eq:adapter-mixing}
\end{equation}
The logits are initialized to zero, assigning equal weight to the two branches at the start of training.
They are then optimized with the local task loss and kept private.
This mechanism is client- and layer-specific but input-independent: it learns how strongly each client should rely on shared and private updates at each adapted projection, without introducing an example-level routing network.
The fusion in Eq.~\eqref{eq:dual-adapter-layer} is performed within each target layer by weighting and summing the shared and private adapter outputs, rather than combining the final predictions of two separate models.

\subsection{Federated Training and Aggregation}
\label{sec:method-training}

Training alternates between joint local adaptation and server aggregation.
At communication round $t$, the server selects a set of clients $\mathcal{S}_t$ and broadcasts the current shared LoRA parameters $\Theta^{\mathrm{sh},t}$.
Each selected client replaces only this synchronized component; its private RandLoRA parameters and mixing logits are carried over from its previous local state.
The client then jointly updates the received shared parameters, private RandLoRA scaling variables, and mixing logits for $E$ local epochs.

After local training, client $i$ uploads $\Theta_i^{\mathrm{sh},t+1}$ but withholds $\Theta_i^{\mathrm{pr},t+1}$ and $z_i^{t+1}$.
With
\begin{equation}
    p_i^t
    = \frac{N_i}{\sum_{k\in\mathcal{S}_t}N_k},
    \label{eq:aggregation-weight}
\end{equation}
the server performs sample-weighted aggregation:
\begin{equation}
    \Theta^{\mathrm{sh},t+1}
    = \sum_{i\in\mathcal{S}_t}p_i^t
      \Theta_i^{\mathrm{sh},t+1}.
    \label{eq:fedlafp-aggregation}
\end{equation}
For the shared LoRA branch, Eq.~\eqref{eq:fedlafp-aggregation} averages the $A^{\mathrm{sh}}$ and $B^{\mathrm{sh}}$ factors separately, corresponding to the default FedAvg realization used by FedLAFP. 
Since the private RandLoRA parameters and fusion logits are never communicated, their maintenance is decoupled from server-side aggregation of the shared branch. 
The FedAvg step in Eq.~\eqref{eq:fedlafp-aggregation} can therefore be replaced or augmented by federated optimization techniques that correct factor-wise aggregation errors~\cite{singhal2025fedex}, selectively aggregate LoRA components~\cite{guo2025selective}, or add proximal regularization~\cite{li2020federated}, without modifying the private branch or the fusion mechanism.

The complete training algorithm and convergence analysis are provided in the supplementary material.

\subsubsection{Personalized inference.} After federated training, client $i$ performs inference using the final shared LoRA parameters together with its locally retained private RandLoRA parameters and mixing logits, following Eq.~\eqref{eq:dual-adapter-layer}. 

\begin{table*}[t]
    \centering
    \begin{tabular}{lccccc}
        \toprule
        \textbf{Method} & \textbf{DTD} & \textbf{Oxford Pets} & \textbf{SUN397} & \textbf{UCF101} & \textbf{Average} \\
        \midrule
        Local-only LoRA & $67.42 \pm 0.30$ & $90.81 \pm 0.17$ & $78.97 \pm 0.03$ & $83.57 \pm 0.22$ & $80.19$ \\
        Local-only RandLoRA & $67.99 \pm 0.36$ & $91.11 \pm 0.22$ & $78.99 \pm 0.07$ & $83.70 \pm 0.20$ & $80.45$ \\
        FedLoRA~{\scriptsize [ICASSP'24]} & $74.59 \pm 0.41$ & $95.47 \pm 0.07$ & $83.02 \pm 0.04$ & $87.81 \pm 0.11$ & $85.22$ \\
        FedRandLoRA & $75.00 \pm 0.31$ & $95.43 \pm 0.07$ & $83.89 \pm 0.07$ & $88.20 \pm 0.09$ & $85.63$ \\
        \midrule
        FFA-LoRA~{\scriptsize [ICLR'24]} & $74.35 \pm 0.36$ & $95.48 \pm 0.08$ & $82.86 \pm 0.02$ & $87.86 \pm 0.23$ & $85.14$ \\
        FedEx-LoRA~{\scriptsize [ACL'25]} & $74.47 \pm 0.12$ & $95.45 \pm 0.09$ & $82.95 \pm 0.01$ & $87.73 \pm 0.03$ & $85.15$ \\
        FedSA-LoRA~{\scriptsize [ICLR'25]} & $74.55 \pm 0.38$ & $95.38 \pm 0.18$ & $83.46 \pm 0.07$ & $87.97 \pm 0.26$ & $85.34$ \\
        FedALT~{\scriptsize [AAAI'26]} & $74.59 \pm 0.41$ & $95.40 \pm 0.12$ & $83.02 \pm 0.07$ & $87.71 \pm 0.13$ & $85.18$ \\
        \midrule
        \textbf{FedLAFP (Ours)} & $\mathbf{78.62 \pm 0.03}$ & $\mathbf{95.96 \pm 0.03}$ & $\mathbf{84.00 \pm 0.07}$ & $\mathbf{89.12 \pm 0.08}$ & $\mathbf{86.93}$ \\
        \bottomrule
    \end{tabular}
    \caption{Comparison with local-only and federated LoRA baselines on four visual recognition benchmarks.
    Accuracy (\%) is reported as mean $\pm$ standard deviation over three runs.
    The best result in each column is shown in bold.}
    \label{tab:main-results}
\end{table*}

\section{Experiments}
\label{sec:experiments}

\subsection{Experimental Setup}
\label{sec:experimental-setup}

\subsubsection{Datasets and backbone.}
We evaluate on four visual recognition benchmarks spanning complementary tasks: DTD for texture recognition~\cite{cimpoi2014describing}, Oxford Pets for fine-grained pet recognition~\cite{parkhi2012cats}, SUN397 for scene recognition~\cite{xiao2010sun}, and UCF101 for human-action recognition~\cite{soomro2012ucf101}.
To construct a few-shot federated adaptation setting, we globally sample 16 examples per class from the predefined training split, while retaining the full predefined test split for evaluation.
To simulate heterogeneity, a Dirichlet proportion vector ($\alpha=0.1$) is drawn for each class and used to partition that class’s training and test examples among 12 clients.
We use a pre-trained ViT-B/16~\cite{dosovitskiy2021an}.
The Transformer backbone remains frozen, and adaptation is restricted to the query and value projections in all self-attention blocks.
We additionally evaluate RoBERTa-Large~\cite{liu2019roberta} on MRPC and SST-2 from GLUE~\cite{wang2018glue} to assess the transferability of FedLAFP beyond visual adaptation; the detailed setup for these language experiments is provided in the supplementary material.

\subsubsection{Baselines.}
\emph{Local-only LoRA} and \emph{Local-only RandLoRA} train an independent adapter on each client without communication, providing local-training references for low-rank and full-rank-capable adaptation, respectively.
\emph{FedLoRA} applies the same sample-weighted aggregation of trainable LoRA parameters as FedIT~\cite{zhang2024towards}, while \emph{FedRandLoRA} analogously aggregates the learned RandLoRA scaling parameters.
\emph{FFA-LoRA}~\cite{sunimproving} freezes the randomly initialized LoRA $A$ factors and trains only the $B$ factors.
\emph{FedEx-LoRA}~\cite{singhal2025fedex} enables exact aggregation by correcting the mismatch introduced by separately averaging the LoRA factors.
\emph{FedSA-LoRA}~\cite{guo2025selective} aggregates the LoRA $A$ factors while retaining the $B$ factors locally.
\emph{FedALT}~\cite{bian2026fedalt} combines an individual LoRA with a Rest-of-World LoRA through an adaptive mixer.
All baselines use the same pre-trained backbone, client partitions, and query--value target modules.

\subsubsection{Implementation details.}
Unless otherwise specified, we use the same training protocol for FedLAFP and all baselines.
All federated methods are trained for $50$ communication rounds, with each participating client performing three local epochs per round using a batch size of $64$.
The local-only baselines are trained for an equivalent number of local epochs.
Local optimization uses SGD with learning rate $0.1$.
For FedLAFP, the shared LoRA rank is set to $r_L=4$, and the private RandLoRA rank is set to $r_R=128$.
We choose these ranks to match the numbers of trainable parameters in the two adapter branches as closely as possible, enabling a fair comparison between their parameterizations.
For each baseline, we adjust the adapter rank so that the total number of trainable parameters remains approximately matched across methods.
For all methods, we set each adapter's scaling parameter equal to its rank, yielding a multiplicative adapter scale of $\alpha/r=1$.
At test time, we evaluate each method on every client's local test set.
Personalized methods use the corresponding client-specific model, whereas methods that learn a single global model use that model for all clients.
The accuracy of one run is computed over the union of all client predictions, equivalently as the test-sample-weighted mean of the client accuracies.
We repeat the visual experiment three times and report the mean and standard deviation across runs.

\begin{table}[t]
\centering
\begin{tabular}{lccc}
\toprule
\textbf{Method} & \textbf{MRPC} & \textbf{SST-2} & \textbf{Average} \\
\midrule
FedLoRA~{\scriptsize [ICASSP'24]} & 72.30 & 93.81 & 83.06 \\
FedALT~{\scriptsize [AAAI'26]} & 88.24 & 95.41 & 91.83 \\
\textbf{FedLAFP (Ours)} & \textbf{88.73} & \textbf{95.76} & \textbf{92.25} \\
\bottomrule
\end{tabular}
\caption{Accuracy (\%) on GLUE language tasks using RoBERTa-Large.}
\label{tab:language-results}
\end{table}

\subsection{Performance Comparisons}
\label{sec:main-results}

\subsubsection{Visual benchmarks.} Table~\ref{tab:main-results} compares FedLAFP with local-only and federated LoRA baselines. FedLAFP achieves the best accuracy on all four benchmarks, with an average accuracy of $86.93\%$. 
Its consistent gains across texture, fine-grained pet, scene, and action recognition suggest that the role-aware design is effective across diverse visual recognition tasks.
FedLoRA and FedRandLoRA outperform their corresponding local-only variants by $5.03$ and $5.18$ percentage points on average, respectively, confirming the benefit of cross-client knowledge sharing. FedLAFP further improves over FedLoRA and FedRandLoRA by $1.71$ and $1.30$ percentage points, respectively, suggesting that combining shared aggregation with client-specific adaptation is more effective than fully aggregating a single adapter under heterogeneous client distributions. FedLAFP also outperforms all remaining federated LoRA baselines. The best of these baselines achieves an average accuracy of $85.34\%$, compared with $86.93\%$ for FedLAFP, further demonstrating the effectiveness of the proposed role-aware design.

\subsubsection{Language benchmarks.} Table~\ref{tab:language-results} reports the results on MRPC and SST-2 using RoBERTa-Large. FedLAFP outperforms both FedLoRA and FedALT on the two tasks, exceeding the stronger FedALT baseline by $0.49$ and $0.35$ percentage points on MRPC and SST-2, respectively. These results show that the role-aware design is also effective beyond visual adaptation.

\begin{figure}[t]
    \centering
    \includegraphics[width=0.9\columnwidth]{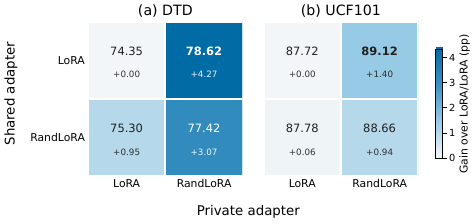}
    \caption{Role-assignment matrices on DTD and UCF101. Rows specify the shared adapter and columns specify the private adapter. Each cell reports accuracy (\%); the smaller annotation and cell color encode the gain over LoRA/LoRA. Bold values denote the best assignment per dataset.}
    \label{fig:adapter-assignment-matrices}
\end{figure}

\subsection{Ablation Study}

\subsubsection{Impact of Adapter Role Assignment.}
\label{sec:adapter-assignment-ablation}

We evaluate all four combinations of LoRA and RandLoRA in the shared and private branches while keeping the adaptive fusion mechanism and training protocol fixed.
Figure~\ref{fig:adapter-assignment-matrices} presents the DTD and UCF101 results as $2\times2$ matrices, with shared adapters along the rows and private adapters along the columns.
In both matrices, the upper-right cell, corresponding to shared LoRA and private RandLoRA, achieves the highest accuracy.
The complete four-dataset results provided in the supplementary material further show that this assignment also performs best on Oxford Pets and SUN397.
Its consistent advantage across all four datasets supports assigning compact LoRA to shared aggregation and expressive RandLoRA to private adaptation.

\begin{table}[t]
    \centering
    \begin{tabular}{lcccc}
        \toprule
        \textbf{Scheme} & \textbf{DTD} & \textbf{Pets} & \textbf{SUN} & \textbf{UCF} \\
        \midrule
        Private $(0,1)$ & $67.99$ & $91.11$ & $78.99$ & $83.70$ \\
        Shared $(1,0)$ & $74.59$ & $95.47$ & $83.02$ & $87.81$ \\
        Equal $(0.5,0.5)$ & $77.48$ & $95.78$ & $83.96$ & $88.82$ \\
        Instance gate & $78.07$ & $95.86$ & $82.49$ & $88.76$ \\
        Client--layer (Ours) & $\mathbf{78.62}$ & $\mathbf{95.96}$ & $\mathbf{84.00}$ & $\mathbf{89.12}$ \\
        \bottomrule
    \end{tabular}
    \caption{Comparison of strategies for fusing the shared and private branches.}
    \label{tab:fusion-weight-ablation}
\end{table}

\begin{figure*}[t]
    \centering
    \includegraphics[width=0.9\textwidth]{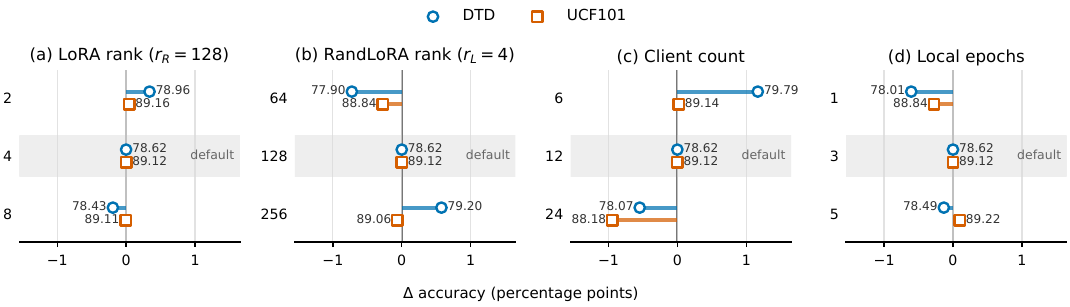}
    \caption{Hyperparameter sensitivity on DTD and UCF101. Each panel varies one hyperparameter while holding the remaining settings fixed. The horizontal axis shows the accuracy change in percentage points relative to the default configuration $(r_L,r_R,K,E)=(4,128,12,3)$; marker labels show absolute accuracy (\%). The shaded row denotes the default value.}
    \label{fig:hyperparameter-sensitivity}
\end{figure*}

\begin{figure}[t]
    \centering
    \includegraphics[width=0.9\columnwidth]{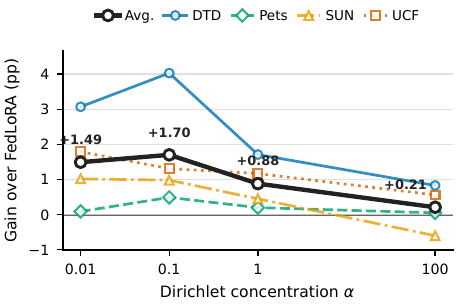}
    \caption{Accuracy gain of FedLAFP over FedLoRA under different Dirichlet concentration parameters. Positive values indicate an improvement over FedLoRA under the same client partition. The black curve shows the mean gain across the four datasets; larger $\alpha$ corresponds to less heterogeneous client data.}
    \label{fig:dirichlet-robustness-gain}
\end{figure}

\subsubsection{Impact of Fusion Weights.}
\label{sec:fusion-weight-ablation}

Table~\ref{tab:fusion-weight-ablation} compares five ways of combining the private and shared branches.
The private-only $(0,1)$ and shared-only $(1,0)$ variants use one branch exclusively, while the equal $(0.5,0.5)$ variant assigns a fixed weight of 0.5 to each branch.
The instance-gate variant uses an auxiliary gating unit to predict input-dependent weights for each sample.
Our client--layer scheme instead learns input-independent weights separately for each client and adapted layer.
The proposed client--layer fusion achieves the highest accuracy on all four benchmarks.
It requires no auxiliary gating unit and introduces only two trainable scalar logits per adapted layer for each client.

\subsection{Sensitivity Analysis}
\label{sec:sensitivity-analysis}

\paragraph{Adapter Ranks.}
Figures~\ref{fig:hyperparameter-sensitivity}(a) and (b) show that the shared LoRA rank has a relatively small effect on accuracy, while the private RandLoRA rank has a somewhat larger effect.
The degree of sensitivity also depends on the dataset, with DTD responding more strongly to rank changes than UCF101.

\paragraph{Number of Clients.}
Figure~\ref{fig:hyperparameter-sensitivity}(c) shows that accuracy generally decreases as the number of clients increases.
With a fixed amount of training data, more clients divide the data into smaller local partitions, making the federated learning task more difficult and reducing accuracy.

\paragraph{Local Epochs.}
As shown in Fig.~\ref{fig:hyperparameter-sensitivity}(d), increasing the number of local epochs from $1$ to $3$ improves accuracy on both datasets.
Further increasing it to $5$ provides little additional benefit, indicating diminishing returns from additional local training.

\paragraph{Data Heterogeneity.}
We vary the Dirichlet concentration parameter over $\alpha\in\{0.01,0.1,1,100\}$, where smaller values produce more heterogeneous client partitions and $\alpha=100$ approximates an IID allocation.
Figure~\ref{fig:dirichlet-robustness-gain} shows that FedLAFP improves over FedLoRA by $1.49$, $1.70$, $0.88$, and $0.21$ percentage points on average for $\alpha=0.01$, $0.1$, $1$, and $100$, respectively; the corresponding absolute accuracies are provided in the supplementary material.
The gain peaks at $\alpha=0.1$: extreme fragmentation at $\alpha=0.01$ limits the common knowledge available to the shared branch, whereas increasingly IID partitions provide less client-specific information for the private branch.
Thus, FedLAFP remains beneficial across heterogeneous and nearly IID settings, with the largest improvement when shared and personalized knowledge are both informative.

\section{Conclusion}
\label{sec:conclusion}

We introduced FedLAFP, a role-aware framework that assigns compact LoRA to shared aggregation and full-rank-capable RandLoRA to private adaptation, with client- and layer-specific fusion.
FedLAFP communicates only the shared LoRA parameters, while the private RandLoRA parameters and fusion logits remain local.
Controlled analysis showed that LoRA produces more aggregation-stable updates, whereas RandLoRA better recovers client-specific residuals.
Across four visual benchmarks, FedLAFP achieved the best accuracy on every dataset and an average of $86.93\%$, exceeding the strongest baseline by $1.30$ percentage points.
Ablation studies and sensitivity analyses further confirmed the value of the asymmetric design under data heterogeneity.
Future work will extend FedLAFP to larger models, investigate realistic resource heterogeneity across clients, incorporate privacy-preserving aggregation, and explore automatic allocation of shared and private adaptation capacity.

\bibliography{aaai2027}

@article{guo2023promptfl,
  title={Promptfl: Let federated participants cooperatively learn prompts instead of models--federated learning in age of foundation model},
  author={Guo, Tao and Guo, Song and Wang, Junxiao and Tang, Xueyang and Xu, Wenchao},
  journal={IEEE Transactions on Mobile Computing},
  volume={23},
  number={5},
  pages={5179--5194},
  year={2023},
  publisher={IEEE}
}

@inproceedings{mcmahan2017communication,
  title={Communication-efficient learning of deep networks from decentralized data},
  author={McMahan, Brendan and Moore, Eider and Ramage, Daniel and Hampson, Seth and y Arcas, Blaise Aguera},
  booktitle={Artificial intelligence and statistics},
  pages={1273--1282},
  year={2017},
  organization={Pmlr}
}

@inproceedings{hu2022lora,
title={Lo{RA}: Low-Rank Adaptation of Large Language Models},
author={Edward J Hu and Yelong Shen and Phillip Wallis and Zeyuan Allen-Zhu and Yuanzhi Li and Shean Wang and Lu Wang and Weizhu Chen},
booktitle={International Conference on Learning Representations},
year={2022},
}

@inproceedings{chen2026fedmerge,
  title={Fedmerge: Federated model merging for personalization},
  author={Chen, Shutong and Zhou, Tianyi and Long, Guodong and Jiang, Jing and Zhang, Chengqi},
  booktitle={Proceedings of the AAAI Conference on Artificial Intelligence},
  volume={40},
  number={24},
  pages={20253--20261},
  year={2026}
}

@inproceedings{bian2026fedalt,
  title={Fedalt: Federated fine-tuning through adaptive local training with rest-of-world lora},
  author={Bian, Jieming and Wang, Lei and Zhang, Letian and Xu, Jie},
  booktitle={Proceedings of the AAAI Conference on Artificial Intelligence},
  volume={40},
  number={24},
  pages={19728--19736},
  year={2026}
}

@inproceedings{
albert2025randlora,
title={RandLo{RA}: Full rank parameter-efficient fine-tuning of large models},
author={Paul Albert and Frederic Z. Zhang and Hemanth Saratchandran and Cristian Rodriguez-Opazo and Anton van den Hengel and Ehsan Abbasnejad},
booktitle={The Thirteenth International Conference on Learning Representations},
year={2025},
}

@article{ding2023parameter,
  title={Parameter-efficient fine-tuning of large-scale pre-trained language models},
  author={Ding, Ning and Qin, Yujia and Yang, Guang and Wei, Fuchao and Yang, Zonghan and Su, Yusheng and Hu, Shengding and Chen, Yulin and Chan, Chi-Min and Chen, Weize and others},
  journal={Nature machine intelligence},
  volume={5},
  number={3},
  pages={220--235},
  year={2023},
  publisher={Nature Publishing Group UK London}
}

@inproceedings{lu2023fedclip,
title={Fedclip: Fast generalization and personalization for clip in federated learning},
author={Wang Lu and Xixu Hu and Jindong Wang and Xing Xie},
booktitle={ICLR 2023 Workshop on Trustworthy and Reliable Large-Scale Machine Learning Models },
year={2023},
}

@article{zhou2022learning,
  title={Learning to prompt for vision-language models},
  author={Zhou, Kaiyang and Yang, Jingkang and Loy, Chen Change and Liu, Ziwei},
  journal={International journal of computer vision},
  volume={130},
  number={9},
  pages={2337--2348},
  year={2022},
  publisher={Springer}
}

@inproceedings{zaken2022bitfit,
  title={Bitfit: Simple parameter-efficient fine-tuning for transformer-based masked language-models},
  author={Zaken, Elad Ben and Goldberg, Yoav and Ravfogel, Shauli},
  booktitle={Proceedings of the 60th Annual Meeting of the Association for Computational Linguistics (Volume 2: Short Papers)},
  pages={1--9},
  year={2022}
}

@inproceedings{zhang2023adalora,
title={Adaptive Budget Allocation for Parameter-Efficient Fine-Tuning },
author={Qingru Zhang and Minshuo Chen and Alexander Bukharin and Pengcheng He and Yu Cheng and Weizhu Chen and Tuo Zhao},
booktitle={The Eleventh International Conference on Learning Representations },
year={2023},
}

@inproceedings{liu2024dora,
  title={Dora: Weight-decomposed low-rank adaptation},
  author={Liu, Shih-Yang and Wang, Chien-Yi and Yin, Hongxu and Molchanov, Pavlo and Wang, Yu-Chiang Frank and Cheng, Kwang-Ting and Chen, Min-Hung},
  booktitle={Forty-first International Conference on Machine Learning},
  year={2024}
}

@inproceedings{kopiczko2024vera,
title={Ve{RA}: Vector-based Random Matrix Adaptation},
author={Dawid Jan Kopiczko and Tijmen Blankevoort and Yuki M Asano},
booktitle={The Twelfth International Conference on Learning Representations},
year={2024},
}

@article{fallah2020personalized,
  title={Personalized federated learning with theoretical guarantees: A model-agnostic meta-learning approach},
  author={Fallah, Alireza and Mokhtari, Aryan and Ozdaglar, Asuman},
  journal={Advances in neural information processing systems},
  volume={33},
  pages={3557--3568},
  year={2020}
}

@inproceedings{oh2022fedbabu,
title={Fed{BABU}: Toward Enhanced Representation for Federated Image Classification},
author={Jaehoon Oh and SangMook Kim and Se-Young Yun},
booktitle={International Conference on Learning Representations},
year={2022},
}

@inproceedings{collins2021exploiting,
  title={Exploiting shared representations for personalized federated learning},
  author={Collins, Liam and Hassani, Hamed and Mokhtari, Aryan and Shakkottai, Sanjay},
  booktitle={International conference on machine learning},
  pages={2089--2099},
  year={2021},
  organization={PMLR}
}

@inproceedings{
chen2022on,
title={On Bridging Generic and Personalized Federated Learning for Image Classification},
author={Hong-You Chen and Wei-Lun Chao},
booktitle={International Conference on Learning Representations},
year={2022},
}

@inproceedings{sunimproving,
  title={Improving LoRA in Privacy-preserving Federated Learning},
  author={Sun, Youbang and Li, Zitao and Li, Yaliang and Ding, Bolin},
  booktitle={The Twelfth International Conference on Learning Representations},
  year={2024}
}

@inproceedings{singhal2025fedex,
  title={FedEx-LoRA: Exact aggregation for federated and efficient fine-tuning of large language models},
  author={Singhal, Raghav and Ponkshe, Kaustubh and Vepakomma, Praneeth},
  booktitle={Proceedings of the 63rd Annual Meeting of the Association for Computational Linguistics (Volume 1: Long Papers)},
  pages={1316--1336},
  year={2025}
}

@inproceedings{guo2025selective,
title={Selective Aggregation for Low-Rank Adaptation in Federated Learning},
author={Pengxin Guo and Shuang Zeng and Yanran Wang and Huijie Fan and Feifei Wang and Liangqiong Qu},
booktitle={The Thirteenth International Conference on Learning Representations},
year={2025}
}

@inproceedings{hefedpissa,
  title={FedPissa: Towards Federated Personalized Adaptation of Foundation Models via LoRA Subspace Mapping},
  author={He, Wenwen and Huang, Wenke and Liu, Yi and Liang, Jian and Li, Xirui and Pang, Guansong and Ye, Mang},
  booktitle={Forty-third International Conference on Machine Learning},
  year={2026}
}

@inproceedings{yang2024dual,
title={Dual-Personalizing Adapter for Federated Foundation Models},
author={Yiyuan Yang and Guodong Long and Tao Shen and Jing Jiang and Michael Blumenstein},
booktitle={The Thirty-eighth Annual Conference on Neural Information Processing Systems},
year={2024},
}

@article{lu2024fdlora,
  title={Fdlora: Personalized federated learning of large language model via dual lora tuning},
  author={Lu, Yao and Qi, Jiaxing and Luan, Zhongzhi and Huang, Shaohan and Fung, Carol and Yang, Hailong and Qian, Depei},
  journal={arXiv preprint arXiv:2406.07925},
  year={2024}
}

@article{li2020federated,
  title={Federated optimization in heterogeneous networks},
  author={Li, Tian and Sahu, Anit Kumar and Zaheer, Manzil and Sanjabi, Maziar and Talwalkar, Ameet and Smith, Virginia},
  journal={Proceedings of Machine learning and systems},
  volume={2},
  pages={429--450},
  year={2020}
}

@inproceedings{cimpoi2014describing,
  title={Describing textures in the wild},
  author={Cimpoi, Mircea and Maji, Subhransu and Kokkinos, Iasonas and Mohamed, Sammy and Vedaldi, Andrea},
  booktitle={Proceedings of the IEEE conference on computer vision and pattern recognition},
  pages={3606--3613},
  year={2014}
}

@inproceedings{parkhi2012cats,
  title={Cats and dogs},
  author={Parkhi, Omkar M and Vedaldi, Andrea and Zisserman, Andrew and Jawahar, CV},
  booktitle={2012 IEEE conference on computer vision and pattern recognition},
  pages={3498--3505},
  year={2012},
  organization={IEEE}
}

@inproceedings{xiao2010sun,
  title={Sun database: Large-scale scene recognition from abbey to zoo},
  author={Xiao, Jianxiong and Hays, James and Ehinger, Krista A and Oliva, Aude and Torralba, Antonio},
  booktitle={2010 IEEE computer society conference on computer vision and pattern recognition},
  pages={3485--3492},
  year={2010},
  organization={IEEE}
}

@article{soomro2012ucf101,
  title={Ucf101: A dataset of 101 human actions classes from videos in the wild},
  author={Soomro, Khurram and Zamir, Amir Roshan and Shah, Mubarak},
  journal={arXiv preprint arXiv:1212.0402},
  year={2012}
}

@inproceedings{wang2018glue,
title={{GLUE}: A Multi-Task Benchmark and Analysis Platform for Natural Language Understanding},
author={Alex Wang and Amanpreet Singh and Julian Michael and Felix Hill and Omer Levy and Samuel R. Bowman},
booktitle={International Conference on Learning Representations},
year={2019},
}

@inproceedings{zhang2024towards,
  title={Towards building the federatedgpt: Federated instruction tuning},
  author={Zhang, Jianyi and Vahidian, Saeed and Kuo, Martin and Li, Chunyuan and Zhang, Ruiyi and Yu, Tong and Wang, Guoyin and Chen, Yiran},
  booktitle={ICASSP 2024-2024 IEEE international conference on acoustics, speech and signal processing (ICASSP)},
  pages={6915--6919},
  year={2024},
  organization={IEEE}
}

@inproceedings{fu2023effectiveness,
  title={On the effectiveness of parameter-efficient fine-tuning},
  author={Fu, Zihao and Yang, Haoran and So, Anthony Man-Cho and Lam, Wai and Bing, Lidong and Collier, Nigel},
  booktitle={Proceedings of the AAAI conference on artificial intelligence},
  volume={37},
  number={11},
  pages={12799--12807},
  year={2023}
}

@inproceedings{yang2025federated,
  title     = {Federated Low-Rank Adaptation for Foundation Models: A Survey},
  author    = {Yang, Yiyuan and Long, Guodong and Lu, Qinghua and Zhu, Liming and Jiang, Jing and Zhang, Chengqi},
  booktitle = {Proceedings of the Thirty-Fourth International Joint Conference on Artificial Intelligence, {IJCAI-25}},
  publisher = {International Joint Conferences on Artificial Intelligence Organization},
  pages     = {10779--10787},
  year      = {2025},
  month     = {8},
}

@article{liu2019roberta,
  title={Roberta: A robustly optimized bert pretraining approach},
  author={Liu, Yinhan and Ott, Myle and Goyal, Naman and Du, Jingfei and Joshi, Mandar and Chen, Danqi and Levy, Omer and Lewis, Mike and Zettlemoyer, Luke and Stoyanov, Veselin},
  journal={arXiv preprint arXiv:1907.11692},
  year={2019}
}

@inproceedings{
dosovitskiy2021an,
title={An Image is Worth 16x16 Words: Transformers for Image Recognition at Scale},
author={Alexey Dosovitskiy and Lucas Beyer and Alexander Kolesnikov and Dirk Weissenborn and Xiaohua Zhai and Thomas Unterthiner and Mostafa Dehghani and Matthias Minderer and Georg Heigold and Sylvain Gelly and Jakob Uszkoreit and Neil Houlsby},
booktitle={International Conference on Learning Representations},
year={2021},
}


\end{document}